\documentclass[twoside,11pt]{article}

\usepackage[preprint]{jmlr2e}
\usepackage{amsmath}
\usepackage{nicefrac}
\usepackage{color}
\usepackage{afterpage}

\ShortHeadings{Stabilizing GANs: A Survey}{Wiatrak, Albrecht, Nystrom}
\firstpageno{1}

\begin{document}

\title{Stabilizing Generative Adversarial Networks: A Survey}

\author{\name Maciej Wiatrak \email macwiatrak@gmail.com \\
        \addr School of Informatics \\ University of Edinburgh
        \AND
        \name Stefano V. Albrecht \email s.albrecht@ed.ac.uk \\
        \addr School of Informatics \\ University of Edinburgh
        \AND
        \name Andrew Nystrom \email nystrom@google.com \\
        \addr Google Research
}

\editor{}

\maketitle

\begin{abstract}
    Generative Adversarial Networks (GANs) are a type of generative model which have received much attention due to their ability to model complex real-world data. Despite their recent successes, the process of training GANs remains challenging, suffering from instability problems such as non-convergence, vanishing or exploding gradients, and mode collapse. In recent years, a diverse set of approaches have been proposed which focus on stabilizing the GAN training procedure. The purpose of this survey is to provide a comprehensive overview of the GAN training stabilization methods which can be found in the literature. We discuss the advantages and disadvantages of each approach, offer a comparative summary, and conclude with a discussion of open problems.
\end{abstract}


\section{Introduction}
\label{sec:intro}

Generative Adversarial Networks (GANs) \citep{Goodfellow2014a} are a type of generative model which have received much attention in recent years due to their ability to model complex real-world data. Comprised of two models, a generator ($G$) and a discriminator ($D$) both typically represented as neural networks, the two models are trained in a process of adversarial optimization. GANs have been shown to achieve state-of-the-art results in the generation of images \citep{Odena2017,Radford2016,Salimans2016,Zhu2017}, natural language \citep{Yu2016}, time-series synthesis \citep{Donahue2019, Esteban2017}, and other domains \citep{Creswell2018, Wang2017}.

Although GANs achieve state-of-the-art results in many domains, they are known to be notoriously difficult to train, suffering significantly from instability problems \citep{Arjovsky2017, Arjovsky2017a, Salimans2016, Goodfellow2016}. Common failures include mode collapse in which the generator maps different inputs to the same class of outputs which results in highly non-diversified samples; non-convergence due to oscillatory and diverging behaviors during the training of both generator and discriminator; and vanishing or exploding gradients leading to slow learning or a complete stop of the learning process \citep{Goodfellow2016}. 

Recent years have seen a plethora of proposed methods to stailize GAN training, including modified network architectures, loss functions, convergence-aiding training strategies, and other methods \citep{Wang2019, Arjovsky2017, Poole2016, Salimans2016}. Despite the significant progress made, to date none of these methods fully solve the instability problem. This instability poses a significant obstacle to the wider adoption of GANs as an off-the-shelf method, because significant manual parameter tuning is typically required to make the method work \citep{Lucic2018}.

The purpose of this article is to provide a comprehensive survey of GAN stabilization methods and to highlight open problems for the progress of the field. We describe the GAN stability problems, identify and categorize existing stabilization methods into five different categories, and include a comparative summary of methods. While each category is presented separately, it should be noted that many of the proposed methods are complementary and can potentially be jointly applied. 

After discussing related surveys in Section~\ref{sec:relwork}, the reader is introduced to the concept of GANs and instability problems in Section~\ref{sec:gans}. Section~\ref{sec:gan-variants} surveys different GAN variants, outlining strengths and weaknesses of each of the approaches. Section~\ref{sec:trainstab} describes widely employed heuristics for GAN stabilization. Finally, Section~\ref{sec:openprob} outlines open problems which provide useful directions for future research.

    \subsection{Related Surveys}
    \label{sec:relwork}

Most surveys devoted to the field of GANs offer a synopsis of the whole field, outlining key concepts, methods and applications of GANs, albeit without an in-detail description of the GAN training procedure and instability problems \citep{Hong2019, Creswell2018, Wang2017}. \citet{Lucic2018} conduct a large-scale empirical study which focused on the performance of a variety of GANs. They demonstrated a need for more consistent and objective evaluation of GANs, as well as for constructing variants less sensitive to hyperparameter tuning. \citet{Wang2019} provide a descriptive, limited overview of the GAN taxonomy, focusing solely on the architecture and loss function variants. \citet{Hong2019} provide an overview of the current GAN landscape, but omit a number of aspects crucial for stabilization of GAN training. 

In contrast to the surveys mentioned above, a number of studies focus solely on GAN training issues. This includes the work of \citet{Arjovsky2017a} who provide a comprehensive outline of the GAN training issues and the underlying theory. Furthermore, a plethora of work deals with practical considerations \citep{Zhang2018c, Salimans2016, Huszar2016} as well as theoretical aspects \citep{Thanh-Tung2019, Risteski2018, Fedus2018, Mescheder2018, Heusel2017, Arora2017, Mescheder2017a} towards training GANs. A concise practical study by \citet{Kurach2018} investigates the impact of a few chosen loss functions, architecture, normalization and regularization schemes. They explore individual and joint performance, as well as sensitivity to hyperparameter tuning.

Our survey provides a comprehensive summary of current GAN techniques and an in-depth study on GAN training problems, providing an overview of the current state of the GAN research landscape with a specific focus on the problem of GAN training stability. It also aims to describe and categorize GAN variants into a holistic taxonomy, and includes an overview of previously overlooked game theory and gradient-based approaches, as well as general training methods.

\section{Generative Adversarial Networks}
\label{sec:gans}

The basic idea of GANs is a competitive game between the generator, $G$, and discriminator, $D$. The generator is given a Gaussian or uniform random noise $z \sim p_{z}$, and transforms it in such a way that $G(z)$ resembles a target distribution $p_{r}$. Specifically, the generator is trained to maximize the expected log-probability with which the discriminator classifies $G(z)$ as a real sample, $\mathbb{E}_{z \sim p_{z}}\log[D(G(z))]$. Conversely, the discriminator is provided with a set of unlabeled samples from $G(z)$ and $p_{r}$, and is trained to distinguish between the generated (fake) and real samples. Formally, the discriminator is trained to maximize the expected log-probability $\mathbb{E}_{x \sim p_{r}}\log[D(x)]$ while simultaneously minimizing $\mathbb{E}_{z \sim p_{z}}\log[D(G(z))]$. This competition between $G$ and $D$ can be formulated as a two-player zero-sum game in which the following loss function is maximized by the discriminator and minimized by the generator:
\begin{equation}
L(D,G) = \mathbb{E}_{x\sim p_{r}}\log[D(x)] + \mathbb{E}_{z \sim p_{z}}\log[1-D(G(x))]
\end{equation}

In their original publication, \citet{Goodfellow2014a} proved the existence of a unique solution at $D=\frac{1}{2}$, which since GANs are a zero-sum two-player game, corresponds to a Nash equilibrium (NE) in which neither player can improve their cost unilaterally \citep{Nash1951}. In practice, however, GANs have been shown to struggle to reach this NE \citep{Fedus2018,Salimans2016}. Furthermore, the evaluation and benchmarking of GANs has been hampered by the lack of a single universal and comprehensive performance measure.

    \subsection{Instability Problems in GAN Training}
    \label{sec:gan-probs}

Despite the proof of existence of the unique equilibrium in the GAN game, the training dynamics of GANs have proven to be notoriously unstable \citep{Arjovsky2017}. We now discuss the main sources of instability.

In their original formulation, \citet{Goodfellow2014a} have shown that GANs optimize the Jensen-Shannon divergence (JSD)
\begin{equation}
    \text{JSD}(\mathbb{P}_{r}||\mathbb{P}_{g}) = \frac{1}{2}\text{KL}(\mathbb{P}_{r}||\mathbb{P}_{A}) + \frac{1}{2}\text{KL}(\mathbb{P}_{g}||\mathbb{P}_{A})
\end{equation}
where KL denotes the Kullback-Leibler divergence and $\mathbb{P}_{A}$ is defined as the ``average'' distribution $\mathbb{P}_{A} = \frac{P_{r}+P_{g}}{2}$. Using the definition of \text{JSD} and $\mathbb{P}_{A}$, the optimal discriminator $D^{*}$ is
\begin{equation}
    D^* = \frac{P_{r}(x)}{P_{r}(x) + P_{g}(x)}.
\end{equation}
Incorporating $D^*$ into the discriminator's loss as a function of $\theta$, it can be shown that at optimality, the discriminator is minimizing the following equation
\begin{equation}
    L(D^{*}, g_{\theta}) = \text{2JSD}(\mathbb{P}_{r}||\mathbb{P}_{g}) - 2\log2.
\end{equation}

The concept of an optimal discriminator is particularly important because its optimality would ensure that the discriminator produces meaningful feedback to the generator. Therefore to achieve this, it is reasonable to assume that a training procedure would include training the discriminator to optimality for each iteration of the generator. Nevertheless, extensive practical \citep{Salimans2016} and theoretical \citep{Arjovsky2017} studies have shown that training the discriminator to optimality works only in theory. The following sub-sections detail the major causes of this instability.

        \subsubsection{Convergence}

Although the existence of a global NE has been proven \citep{Goodfellow2014a}, arriving at this equilibrium is not straightforward. The proof of convergence described above does not hold in practice due to the fact that the generator and discriminator are modeled as neural networks; thus, the optimization procedure operates in the parameter space of these networks, instead of directly learning the probability density function. Furthermore, the non-convex-concave character of the game makes it especially hard for the gradient descent ascent (GDA) algorithm to converge, often resulting in diverging, oscillating, or cyclic behavior \citep{Mertikopoulos2016, Salimans2016, Goodfellow2014}, and proneness to convergence to a local NE within the parameter space of the neural networks. Furthermore, the local NE can be arbitrarily far from the global NE, thus preventing the discriminator from being optimal. Finally, from a practical standpoint, training the discriminator to optimality for every generator iteration is highly computationally expensive.

        \subsubsection{Vanishing or Exploding Gradients}

According to the theory laid out at the beginning of this section, a discriminator close to optimality would produce meaningful feedback to the generator, thus resulting in an improved generator. However, in practice the cost of the well trained discriminator is not $\text{2JSD}(\mathbb{P}_{r}||\mathbb{P}_{g}) - 2\log2$, but rather approaches $0$. Such a highly accurate discriminator, where $D(x)=1$, $\forall x \in p_{r}$ and $D(G(z))=0$, $\forall G(z) \in p_{z}$ squashes the loss function to 0, producing gradients close to zero, which provides little feedback to the generator, slowing or completely stopping the learning. Hence, as the discriminator gets better at discerning between real and generated samples, the updates to the generator get consistently worse. One of the reasons behind the discriminator easily overpowering the generator is that the supports of real and generated distributions often lie on low dimensional manifolds, making it possible for a perfect discriminator to exist. In such a scenario of a perfect discriminator, its gradients will approach zero on almost every data point, hence providing no feedback to the generator. To avoid the vanishing gradients problem, \citet{Goodfellow2014a} proposed to use an alternative loss for the generator, namely $-\log D(G(z))$, which is widely used in practice and referred to as non-saturating (NS) loss. Although this loss modification alleviates the vanishing gradients problem, change to the loss has not completely solved the problem, in turn contributing to more unstable and oscillating training.

        \subsubsection{Mode Collapse}

One of the most common failures in GANs, \textit{mode collapse} happens when the generator maps multiple distinct inputs to the same output, which means that the generator produces samples of low diversity \citep{Huszar2016, Theis2015}. This is a problem for GAN training, since a generator without the ability to produce diverse samples is not useful for training purposes. Discovering and tackling mode collapse is a difficult problem, since the cause of mode collapse is rooted deeply in the concept of GANs. \citet{Metz2017} argued that with the generator greedily optimizing its loss function in order to ``fool'' the discriminator, each generator update is a partial collapse towards mode collapse. An intuitive solution to mode collapse is a well trained discriminator capable of robustly identifying the symptoms of mode collapse. However, as discussed above, this runs counter to the vanishing or exploding gradients problem. A crucial aspect in addressing mode collapse is a discriminator with high generalization capabilities \citep{Arora2017, Thanh-Tung2019}

All of the above problems contribute to the instability of the GAN training process. These problems are often highly linked with each other and addressing them often takes character of a trade-off. A crucial challenge is the accurate estimation of the generalization capabilities of the discriminator, which in turn would make it possible to properly evaluate the generator. A final obstacle is lack of a comprehensive evaluation metric. At the current state, a number of evaluation metrics have been proposed, such as \textit{Inception} or \textit{Frechet Inception Distance} (FID). Nevertheless, none of them offer a unique solution \citep{Grnarova2018,Lucic2018}.

\section{GAN Variants and Stabilization Techniques}
\label{sec:gan-variants}

\begin{figure}[t]
    \vspace{-1cm}
    \centering
    \includegraphics[angle=90,height=1.00\textheight]{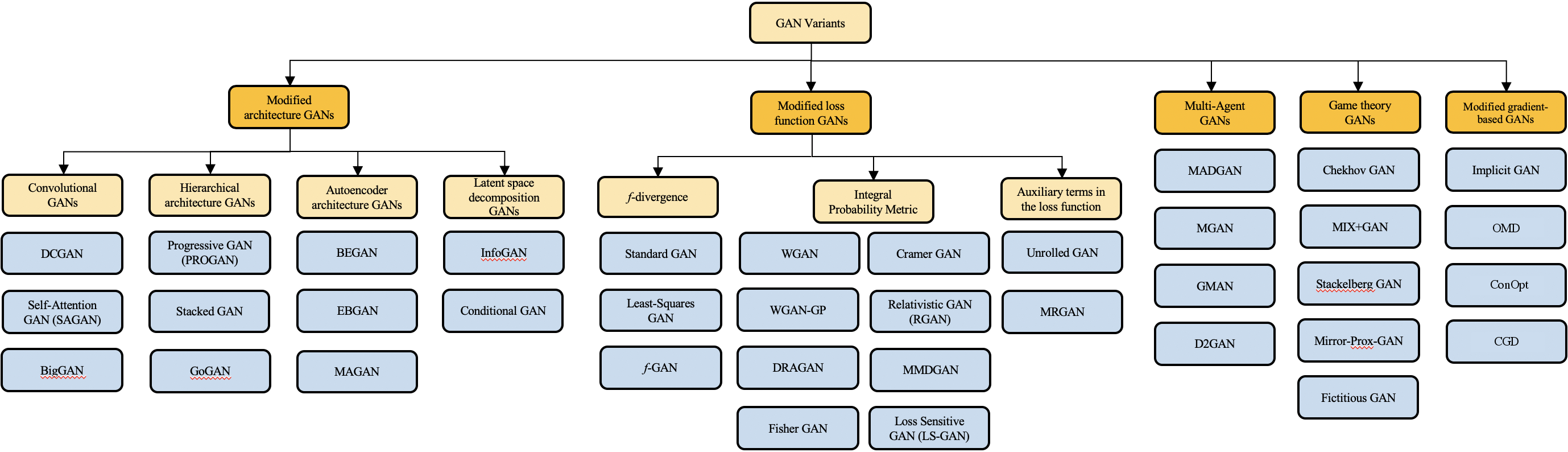}
    \caption{Taxonomy of GAN methods.}
    \label{fig:taxonomy}
\end{figure}

To categorize different methods focused on improving GAN training, the following sections divide the literature into five approaches (cf. Figure~\ref{fig:taxonomy}): architecture, loss function, game theory, multi-agent, and gradient-based. We summarize the key concepts of each approach and discuss their overall advantages and disadvantages. 

\afterpage{\clearpage}

    \subsection{Modified Architectures}

The most extensively studied category is architecture variant GANs. This category covers methods in which the proposed improvements involve networks or training structure while keeping the loss function largely intact from the previously proposed methods. The majority of the architecture-based variants have been established for a specific application \citep{Creswell2018,Hitawala2018,Zhu2017, Yu2016}. Here, we focus on the more general approach, which has shown to improve stability of the training procedure in GANs.

        \subsubsection{Convolutional Architectures}

One of the most popular and successful architecture-variant GANs are inspired by computer vision techniques, specifically Convolutional Neural Networks (CNNs) \citep{Krizhevsky2012}. An example which has been shown to significantly improve the quality of generated images is the Deep Convolutional GAN (DCGAN) \citep{Radford2016}. DCGAN applies a series of convolutional operations including a spatial up-sampling operation to improve the generator. Having shown to enhance the quality of the produced samples solely through architectural changes, DCGAN is often treated as a baseline for modeling other GAN variants \citep{Hong2019}. A further improvement was proposed by \citet{Zhang2018a} in Self-Attention GAN (SAGAN), who incorporated a self-attention mechanism into the design of both networks in order to capture local and global dependencies of the target distribution. Finally, BigGAN \citep{Brock2018} builds on SAGAN and introduces a number of incremental improvements. In particular, both the batch size and number of model parameters are increased, contributing to BigGAN obtaining state-of-the-art results on the ImageNet.

        \subsubsection{Hierarchical Architectures}

A difficult problem in GAN training is the discriminator overpowering the generator in the early stages of the game, thus failing to properly train the generator. This is particularly challenging for generating high resolution images. A heuristic approach designed to tackle this problem is to allow the generator to make a number of updates per one discriminator update \citep{Goodfellow2016}. Nevertheless, this technique has been shown to be limited and often leading to other instability problems. Progressive GAN (PROGAN) \citep{Karras2018} addresses this problem by training the networks in multiple phases, starting from a low-resolution image and minimal network architecture, progressively expanding both networks. Additionally, PROGAN proposes a number of training techniques improving stability for image generation, including pixel normalization and mini-batch standard deviation.

Another hierarchical approach is the simultaneous training of multiple GAN pairs. Stacked GAN (SGAN) \citep{Huang} stacks pairs of generator and discriminator, each consisting of an encoder and decoder on top of each pair. Apart from calculating standard adversarial loss, at each stack the encoder and decoder calculate the discrepancy between generated and real image (referred to as conditional loss) as well as entropy loss, which prevents mode collapse. All three losses are then combined into a composite loss and used in the joint training of all stacks. Similar to Stacked GAN, GoGAN \citep{Juefei-Xu2017} trains pairs of WGANs \citep{Arjovsky2017} (see Section~\ref{sec:intprob}), but uses margin based Wasserstein distance, instead of standard Wasserstein distance. This makes it possible for the discriminator to focus on samples with smaller margin and progressively reduce the margin at each stage of the game. Therefore, steadily decreasing the discrepancy between the generated and target distribution.

        \subsubsection{Autoencoder Architectures}

Another approach designed to prevent the discriminator from optimizing greedily (which may overpower the generator at early stages of the training) is the adoption of an autoencoder architecture for the discriminator. Instead of outputting a probability to distinguish between generated (fake) and real samples, the discriminator assigns an energy value to given samples. The energy model is attained by feeding a sample (generated or real) to a discriminator consisting of an encoder and a decoder, calculating its reconstruction cost, and training it to output high energy level for generated and low for real samples. Intuitively, the autoencoder architecture takes into account the complexity of the image. Furthermore, as the autoencoder is trained to focus on reconstructing the most important features (i.e. near the data manifold), it provides feedback on these features for the generator. The autoencoder architecture was initially proposed by \citet{Zhao2017}, with the loss function being the energy level. Additionally, in order to stabilize training the discriminator ignored the sample outliers with high reconstruction error. The margin restriction was later relaxed by \citet{Wang2017a} in MAGAN, which argued that a fixed margin has a negative influence on the dynamics of the GAN. More complex Boundary Equilibrium GAN (BEGAN) \citep{Berthelot2017} uses Wasserstein distance to match the distributions between reconstructed and generated or real examples. Moreover, it adds a hyperparameter $\gamma = \frac{\mathbb{E}[L(G(z))]}{\mathbb{E}[L(x)]}$ to balance the training between the generator and the discriminator.

        \subsubsection{Latent Space Decomposition Architectures}

One of the reasons behind the instability of training is the low-dimensional support of generated and target distributions, which can lead to the greedy optimization of the discriminator \citep{Arjovsky2017a}. While the dimensions of $p_{r}$ appear to be high, in reality they are usually concentrated in lower dimensional manifolds, which often mirror the most important features of the distribution (for example, a feature of a human face on an image). Such lower dimensional manifolds establish a large number of restrictions for the generator, often making it unattainable for the generator to follow them. This contributes to the disjointedness of generated and real distribution, thus making it easier for the discriminator to perfectly discriminate between the samples.

Latent space decomposition methods aim at providing additional information to the input latent vector $z$, enabling it to disentangle the relevant features from the latent space. InfoGAN \citep{Chen2016} adopts an unsupervised method which supplies a latent variable $c$ capturing the meaningful features of the real sample on top of the standard input noise vector $z$. InfoGAN then maximizes the mutual information $I(c, G(z,c)$ between $z$ and $c$, transforming the generated sample with $G(z, c)$. Due to intractability of estimating the posterior $p(c|x)$, the algorithm adopts a variational approach maximizing a lower bound of $I(c, G(z,c)$, instead of directly calculating it.

In contrast to InfoGAN, Conditional GAN (CGAN) \citep{Odena2017} proposes a supervised method, adding a class label $c$ to a generator and discriminator. This approach has been shown to produce high quality images and prevent mode collapse. However, this is with the strong restriction of possessing a labeled dataset.

        \subsubsection{Summary}

All of the architecture variant GANs and techniques described above have been shown to improve the stability of training in GANs. Nevertheless, the scale at which each of the methods manages to improve the training differs widely. A current strong baseline is BigGAN, which enhances the training in all of the aspects mentioned in the previous section. This improvement, however, is attained at a great computational cost, and improving the training procedure without an increase in complexity remains a challenge. Architecture-variant GANs is a developing and dynamic field, often offering low-hanging-fruits in terms of performance improvements, especially for application-specific GANs. Nevertheless, the non-convex-concave character of the GAN game is a challenge that will require improvements beyond architectural changes.

    \subsection{Modified Loss Functions}

The shortcomings of the standard GAN loss function were already identified by Goodfellow et al. \citep{Goodfellow2014, Goodfellow2014a}, who showed that KL divergence and JS divergence under idealized conditions \citep{Fedus2018} contributes to oscillating and cyclical behavior, especially when the distributions of generated and target data are disjoint. This leads to the non-convergence and vanishing gradients problems. A large number of studies have been devoted to this problem \citep{Wang2019,Lucic2018,Barnett2018,Metz2017,Arjovsky2017a,Poole2016}. In this section, we introduce a number of alternative loss functions employed for GANs and highlight the most important ones.

        \subsubsection{\textit{f}-Divergence}

The KL and JSD divergences discussed in the standard GAN \citep{Goodfellow2014a} belong to an $f$-divergence family, which measures the difference between two given probability distributions. Given distributions $p_{r}$ and $p_{g}$, the $f$-divergence with a function $f$ is defined as:
\begin{equation}
D_{f}(p_{r}\parallel p_{g}) = \int_{\mathcal{X}}p_{g}(x)f\bigg(\frac{p_{r}(x)}{p_{g}(x)}\bigg)dx
\end{equation}
where the generator function $f$ is a convex function and satisfies the condition $f(1) = 0$. In other words, the function $f$ becomes 0 when two distributions are equivalent. The $f$-divergence family can be indirectly calculated by estimating the expectation of the lower bound, which circumvents the intractability problem of unknown probability distributions. \citet{Lollar2002a} discussed the GAN loss function in terms of various divergences and their efficacy under an arbitrary function $f$, proposing f-GAN, which generalizes the GAN loss function through the estimation of various $f$-divergences given $f$. Another f-divergence based GAN is the Least Squares GAN (LSGAN), which tries to remedy the problem of vanishing gradient by replacing the sigmoid cross-entropy loss present in the standard GAN with the least-squares loss, penalizing generated samples far from the decision boundary \citep{Mao2016}. As a divergence, LSGAN has been proven to use the Pearson $\mathcal{X}^{2}$ divergence.

Nevertheless, the $f$-divergence family is subject to a number of limitations. In particular, the growing dimension of data makes it harder to estimate the divergence and the support of the two distributions becomes low dimensional, resulting in the divergence value going to infinity \citep{Sriperumbudur2009}.

        \subsubsection{Integral Probability Metric}
        \label{sec:intprob}

Designed to address the restrictions of the $f$-divergence family, integral probability metrics (IPM) methods are not subject to the data dimension and distribution disjointedness problems, resulting in a consistent distance between the data distributions. Furthermore, in contrast to $f$-divergence in which the discriminator is a binary classifier, the discriminator in an IPM framework is a regression task providing a scalar value and henceforth opening a new avenue for research by treating GANs as an actor-critic problem \citep{Pfau2017}. Defined as a critic function $f$ in a function class $\mathcal{F}$ consisting of real-valued, bounded, measurable functions, an IPM measures the maximal distance between two distributions. Given $p_{r}$ and $p_{g}$ on a compact space $\mathcal{X}$, the IPM metric can be denoted as:
\begin{equation}
d_{\mathcal{F}}(p_{r}, p_{g}) = sup_{f \in \mathcal{F}}\mathbb{E}_{x \sim p_{r}}[f(x)] - \mathbb{E}_{x \sim p_{g}}[f(x)]
\end{equation}

One of the distances which belong to IPM is the \textit{Wasserstein} or Earth Mover's (EM) distance, which is used in Wasserstein GAN (WGAN) \citep{Arjovsky2017} as the new loss measure. Wasserstein distance is defined as:
\begin{equation}
W(p_{r}, p_{g}) = \inf_{\gamma\in\sqcap(p_{r},p_{g})}\mathbb{E}_{(x,y)\sim\gamma}[\parallel x - y \parallel]
\end{equation}

Wasserstein distance has been shown to significantly improve training stability and convergence, dealing particularly well with the distributions support lying on low dimensional manifolds. However, due to the intractability of the infimum term, the critic $f$ needs to be parameterized and weight clipping needs to be applied. A significant drawback is that weight clipping incurs pathological behavior that contributes to slow convergence and not entirely stable training. These problems have been partly addressed by including a gradient penalty instead of weight clipping in the loss function of WGAN-GP \citep{Gulrajani}. Another shortcoming of the WGAN was discussed by \citet{Bellemare2017} who argued that the WGAN incurs biased gradients. Their solution to this problem suggests an energy function similar to Wasserstein distance, but without the biased gradients. Furthermore, the Cramer distance is also linked with the kernel embedded space distance, discussed in the next paragraph on Maximum Mean Discrepancy.

Derived from Generative moment matching networks (GMMN) \citep{Li2015}, Maximum Mean Discrepancy (MMD) distance is an alternative to Wasserstein distance. Continuous and differentiable like the Wasserstein distance, MMD distance has been applied in Maximum Mean Discrepancy GAN (MMDGAN), which can be seen as a combination of GMMNs and GANs \citep{Li2017c}. Essentially, MMD measures the distance between the means of the embedding space of two distributions, using the kernel trick with a Gaussian kernel. In MMDGAN, the Gaussian kernels are replaced by adversarial kernel learning techniques, which is claimed to better represent the feature space. This can be seen as advantageous over WGAN. However, the computational complexity of MMDGAN which increases exponentially with the number of samples is a significant drawback.

A different IPM approach is adopted in Relativistic GAN (RGAN) \citep{Jolicoeur-Martineau2018, Jolicoeur-martineau2019}, in which the authors argue for a general approach to devising new GAN loss functions. This approach estimates the probability that a generated sample is real, taking into account the probability of a corresponding real sample being real. In other words, we measure the distance between the probability of generated sample being real and the probability of real sample being real. It has been shown that the RGAN approach contributes to overcoming the greedy optimization of the discriminator. Furthermore, the technique outlined in RGAN could be potentially expanded to other GANs.

Other IPM approaches include Geometric GAN \citep{Lim2017} which draw on the Support Vector Machine (SVM) algorithm \citep{Cortes1995}, utilizing separating hyperplanes to update the generator and discriminator and encouraging the generator to move toward the separating hyperplane and the discriminator to move away from it. Although proven to improve training stability, the Geometric GAN requires expensive  matrix computation. Loss Sensitive GAN (LS-GAN) \citep{Qi2018} is based on the idea that the loss for a real sample should be smaller than the loss for a generated sample, and aims at minimizing the margins between the two. The reasoning behind LS-GAN is that the non-parametric assumption of the discriminator having infinite capacity \citep{Goodfellow2014a} leads to instability problems, including vanishing gradients, and argues for a more general margin-based approach. Finally, a more stable and computationally efficient approach has been proposed in Fisher GAN \citep{Mroueh2017}, which incorporates a data-dependent constraint into the loss function. Fisher GAN aims at not only reducing the distance between two distributions, but also the in-class variance of them. Furthermore, it avoids adding weight clipping or gradient penalty.

        \subsubsection{Auxiliary Loss Functions}

The modifications in loss function are not limited to devising new divergences from the $f$-divergence or IPM family. An important avenue is providing supplementary terms to the already existing GAN loss function in order to improve the training procedure. Here, we highlight two such methods. The first one is inspired by the recursive reasoning opponent modeling methods \citep{Metz2017}, in which the generators ``unrolls'' possible discriminator reactions and accounts for them by adding a term to the gradient. Given the parameters of the discriminator $\theta_{D}$ and generator $\theta_{G}$, the surrogate loss by unrolling $N$ steps can be defined as:
\begin{equation}
f_{N}(\theta_{D}, \theta_{G}) =  f(\theta_{D}^{N}(\theta_{D}, \theta_{G}), \theta_{G})
\end{equation}

The intuition is that capturing how a discriminator would react given a change in discriminator and feeding it into a generator would allow the generator to react, and prevent mode collapse. A different approach towards preventing mode collapse is offered in Mode Regularized GAN (MRGAN), which uses an encoder $E$ to produce the latent vector $z$ from a real sample $x$, $E(x):x \rightarrow z$ instead of random input noise and alters the loss function by including a term ensuring that different modes are captured in the latent vector $z$.

        \subsubsection{Summary}

Loss function variant GANs are the most heavily studied type of GANs after the architecture type to date. Some of the loss function variant GANs have managed to predominantly overcome the vanishing gradient problem \citep{Arjovsky2017, Jolicoeur-martineau2019, Miyato2018, Mao2016, Qi2018}. Furthermore, in comparison to architecture types, they offer more generalization properties, with methods such as WGAN or SN-GAN being widely applied and offering avenues for further research \citep{Pfau2017}. Additionally, the trade-off between the quality of generated data and computational complexity does not show to be as proportional in comparison to some of the architecture variant GANs such as BigGAN \citep{Wang2019, Brock2018}. However, loss function variant GANs often work only under specific restrictions \citep{Arjovsky2017, Gulrajani}, and proper hyperparameter tuning of architectural design approaches often appear to yield superior results \citep{Lucic2018}. Hence, the stabilization of the training procedure remains a challenge.

    \subsection{Game Theory for GANs}

At the core of all GANs is a competition between generator and discriminator. Game theory GANs focus specifically on that aspect, drawing on the rich literature in the field of two-player competitive games \citep{Freund1999, VonNeumann2007} to aid convergence.

A leading theme in this research area is the computation of mixed strategy Nash equilibria (MNE) \citep{Hsieh2018,Ge2018,Oliehoek2018,Oliehoek2017,Grnarova2017,Arora2017}. In contrast to the pure strategy Nash equilibrium used in standard GAN, which has been proven to guarantee convergence only to a local Nash equilibrium \citep{Via2018,Heusel2017}, under certain circumstances an MNE is always equal to a global NE. A proof for of the existence of the MNE in the GAN game together with a sketch of the suitable algorithm was given by \citet{Arora2017}. Drawing on this work, \citet{Grnarova2017} proposed a convergent algorithm called ChekhovGAN for finding MNE using a regret minimization algorithm \citep{Hazan2017,Freund1999}. However, the ChekhovGAN is only provably convergent under the heavy restrictions that the discriminator is a single-layered network. The authors of \citep{Oliehoek2018,Oliehoek2017} formulated the game as a finite zero-sum game with a finite number of parameters and proposed using a Parallel Nash Memory algorithm \citep{Oliehoek2006} to arrive at the MNE. Their work was further extended by a less restrictive use of Mirror Descent in computing MNEs \citep{Hsieh2018}. A relatively simpler yet effective method to compute MNEs was proposed by \citep{Ge2018}, which trains the model using a mixture of historical models. Some of the game theory inspired methods are also present in other approaches \citep{Kodali2017,Salimans2016}.

        \subsubsection{Summary}

At the current stage, the game theory variant GANs literature is limited, with its proposed methods highly restrictive, thus rarely directly applicable. Nevertheless, work such as \citep{Arora2017} provides excellent theoretical analysis for tackling problems in the training procedure and opens an avenue for further research.

    \subsection{Multi-Agent GANs}

Multi-agent type GANs extend the idea of using a single pair of generator and discriminator to the multi-agent setting and show how employing such methods can provide better convergence properties and prevent mode collapse. The majority of the literature focuses on introducing a larger number of generators \citep{Zhang, Arora2017, Ghosh2017, Nguyen2016} or discriminators \citep{Albuquerque2019, Nguyen2017, Durugkar2017}.

MAD-GAN \citep{Ghosh2017} uses multiple generators and one discriminator not only to identify fake images but also to determine the specific generator which produced the images fed to it. Aimed at preventing mode collapse, MAD-GAN modifies the generator's objective function to encourage different generators to produce different samples. It also changes the discriminator to account for these generators and provide feedback to them accordingly to various modes. An extension of MAD-GAN is MGAN \citep{Nguyen2016} which uses a separate classifier in parallel with discriminator to match the generated output with an appropriate generator. Also providing some theoretical justification for the MGAN enforcing the generators to produce diverse samples, thus prevent mode collapse. Stackelberg GAN \citep{Zhang}, drawing on the concept of \textit{Stackelberg} competition \citep{Sinha2018}, offers an architecture design where multiple generators act as followers to the discriminator (leader), utilizing sampling schemes beyond the mixture model.

At the other side of the spectrum, \citet{Durugkar2017} use many discriminators to boost the learning of the generator, while \citet{Nguyen2017} train exactly two discriminators which separately compute KL and reverse KL divergence to place a fair distribution across the data modes.

MIX+GAN \citep{Arora2017} uses a mixture of multiple generators and discriminators with multiple different parameters to compute mixed strategies. The total reward is then calculated by taking a weighted average of rewards over the pairs of generators and discriminators. An interesting approach is offered by \citet{Ghosh2016} who examine how multi-agent communication can regularize the training of both networks. Previously discussed SGAN \citep{Huang} and GoGAN \citep{Juefei-Xu2017} train GANs in local pairs to establish a global GAN pair trained against all of these pairs at the same time. Similarly, \citet{Im2016} also adopts the concept of multiple GAN pairs, but with discriminators dynamically exchanged during training, thus reducing the effect of coupling which can lead to mode collapse.

        \subsubsection{Summary}

The current literature on multi-agent type GANs mainly revolves around the relatively simple idea of hindering mode collapse through the introduction of a variety of generators and discriminators. Although the results are promising, often the computational costs pose a significant problem. Moreover, it is not yet entirely clear how an appropriate number of generators or discriminators should be chosen. Finally, there remain a variety of methods in multi-agent learning literature which have not been exhausted in terms of stabilizing GAN training \citep{papoudakis2019, Lowe2017, He2016}.

    \subsection{Modified Gradient Optimization}

An interesting research avenue is the development of alternatives to the regular gradient descent ascent (GDA) that are more suitable for general games with multiple losses interacting with each other. Regular GANs trained with GDA have demonstrated diverging and oscillating behaviors during training \citep{Mertikopoulos2016}, and although its convergence has been proven using a two-time scale update rule \citep{Heusel2017}, this requires mild assumptions and can only be guaranteed to arrive at a local Nash equilibrium.  \citet{Daskalakis2017} used Optimistic Mirror Descent (OMD) to approximately converge, but only under a restriction that the game is a two-player bilinear zero-sum game. \citet{Mescheder2018} proposed the \textit{ConOpt} algorithm that adds a penalty for non-convergence. Based on this result, \citet{Balduzzi2018} and later \citet{Letcher2019} decomposed the game into two components: a \textit{potential game} which can be solved by GDA, and a \textit{Hamiltonian game} suitable for solving with \textit{ConOpt} \citep{Mescheder2018}. Finally, \citet{Schafer2019} combined the approaches above and drew on multi-agent methods \citep{papoudakis2019} to create a \textit{Competitive Gradient Descent}, suitable specifically for two-player competitive games. This algorithm was later applied to competitive regularization in GANs \citep{Schafer2019a}, where the generator and discriminator act as agents under local information showing more consistent and stable performance.

        \subsubsection{Summary}

The initial results for GANs with gradient-based variants show improvements in convergence and preventing mode collapse in comparison to GANs employing GDA. Although these results could possibly be achieved with suitable hyperparameter tuning, stabilization of the training is a reasonable contribution. Gradient-based methods remain a largely unexplored field with promising research directions. With GDA's shortcomings in general games, gradient variants provide an additional edge, and importantly, could be used in line with other methods.

    \subsection{Other Modifications}

This section mentions other GAN methods which have no clear correspondence to the previous categories. One is the Evolutionary GAN \citep{Wanga}, which adapts the GAN to the environment through a series of operations on the population of generators and discriminators. AdaGAN \citep{Tolstikhin2017} uses a boosting technique to iteratively address the problem of missing modes. An interesting and deeply theoretical approach was proposed by \citet{Via2018} who provided a proof and a method to escape local equilibria using energy-based potential fields. Another technique for preventing mode collapse includes marginalizing the weights of the generator and discriminator offered in the Bayesian GAN \citep{Saatchi2017}.

    \subsection{Comparative Summary}

To conclude the review of the GAN variants, we outline the advantages and disadvantages as well as the outlook of each of the types. Currently, the approach showing the highest empirical improvement is the architecture type. Nevertheless, this is also due to a large number of publications in the area of modified network architecture relative to other GAN variants, low entry barrier, and some of the designs resulting in meticulous parameter optimization. Furthermore, much of the research of this type lacks rigorous theoretical justification and struggles to alleviate all of the training problems, usually showing improvement in only one or two of them \citep{Wang2019}. At the same time, loss function variant GANs, although often well justified, such as WGAN \citep{Arjovsky2017}, show that their results could also be attained by appropriate hyperparameter optimization of architecture variant GANs such as DCGAN. A comprehensive comparison of GANs \citep{Lucic2018} has demonstrated a small improvement in general metrics by state-of-the-art loss type GANs in comparison to architectures such as DCGAN \citep{Radford2016}, and that it can be more worthwhile to focus on tuning hyperparameters rather than introducing novel methods.

Another survey on convergence in GANs \citep{Fedus2018} highlighted that much of the rigorous theoretical analysis is often not fulfilled in reality; GANs can converge in situations they theoretically should not, and vice versa. It is hypothesized that this is mostly due to the nature of gradient descent ascent. Loss function variant GANs hold a big promise, addressing an obvious issue which is the unsuitability of the standard objective function to the GAN game. Furthermore, they often demonstrate more general improvement in training than architecture type GANs. We believe, however, that a significant breakthrough in loss function GANs would require incorporating an alternative to GDA. This could possibly be addressed by the gradient-based variants, which might have the broadest influence, enhancing all types of GANs by introducing a gradient design relevant to the GAN game. Nevertheless, gradient-based methods are still at a too early stage to form any reliable predictions. Similarly to loss function variants, game theory variants could provide an approach capable of the general improvement in the training procedure, addressing the problem of dealing with the non-convex-concave character of the game. However, current methods in this field work only under strict limitations and cannot compete with architecture and loss function type GANs. It should be noted that this area is also in early stages, with only a handful of studies published. Finally, multi-agent GANs have mostly focused on preventing mode collapse, showing promising but not state-of-the-art results in this problem. A vast area of multi-agent learning including actor-critic methods and opponent modeling \cite{Albrecht2018} remains unexplored.

\section{General Stabilization Heuristics}
\label{sec:trainstab}

A large scale study on GANs \citep{Lucic2018} has shown that optimizing hyperparameters tends to yield superior results compared to applying specific methods such as modified architectures and loss functions. Furthermore, other studies in this area \citep{Fedus2018, Kurach2018} have demonstrated the impact of regularization and normalization techniques on various loss function and architecture GAN variants. In this section, we discuss general heuristics which have demonstrated consistent stabilization improvements in GAN training. These heuristics can potentially be applied to different GAN types.

    \subsection{Regularization}

The most prominent regularization technique in GANs is the gradient penalty, which has been initially applied to WGAN in WGAN-GP \citep{Gulrajani} as a soft penalty constraining the objective function to 1-Lipschitzness. In this study the authors used linear interpolation between real and generated distributions to calculate the gradient norm used for the penalty. The gradient penalty in WGAN was later extended by \citet{Kodali2017}, who looked at the gradient penalty from the perspective of regret minimization. Therefore, using a no-regret type algorithm to constrain the gradient penalty to be calculated only around a data manifold, thus limiting the discriminator functions to the set of linear functions. A different approach was adopted by \citet{Roth2017} who proposed a gradient penalty on the weighted gradient-norm of the discriminator, designed specifically to tackle the issue of non-overlapping support between the generated and target distribution in GANs from \textit{f}-divergence family.

    \subsection{Normalization}

Normalization of the discriminator can stabilize the GAN training procedure from two perspectives. First, it can lead to more stable optimization through inducing the discriminator to produce better quality feedback. Second, it can enrich the representation within the layers of the network of the discriminator.

Optimization-wise, two techniques have been particularly applied to the GAN stability problem. These are batch normalization \citep{Joseph2016} and layer normalization \citep{Ba2016}. Batch normalization, introduced by \citet{denton2015} and popularized by \citet{Radford2016}, has been shown to greatly improve the optimization of neural networks through normalizing the activations of each layer, allowing each layer to learn more stable distribution of inputs. Batch normalization was further extended by \citet{Salimans2016} who proposed virtual batch normalization, which aims to make the input sample less dependent on other samples in the batch, thereby alleviating the mode collapse problem. In contrast to batch normalization, the statistics for layer normalization \citep{Ba2016} are independent of other samples in the batch. However, the normalization of inputs is done across the features, resulting in different samples being normalized differently depending on its features. Layer normalization was initially used in the work of \citet{Gulrajani}.

From a representation standpoint, the most effective method proposed up-to-date has been spectral normalization \citep{Miyato2018}. Introduced initially in the Spectral Normalization GAN (SN-GAN), spectral normalization normalizes the weights of the discriminator, contributing to a more stable training. Spectral normalization constrains the spectral norm of the discriminator's layers, which imposes a Lipschitz condition in a more efficient and adaptable way.

    \subsection{Architectural Features and Training Procedures}

In addition to the regularization and normalisation techniques described above, in recent years the community devised a number of heuristic techniques which have shown to further improve the stability of the training. These include:
\begin{description}
    \item[Feature matching \citep{Salimans2016}] --- Proposed to prevent the generator from overtraining on the given discriminator, feature matching requires the generator to not only maximize the probability of the generated sample being classified as real by the discriminator, but also account for the generated data to match with the target distribution. This is achieved by training the generator on the intermediate layer of the discriminator.
    
    \item[Minibatch discrimination \citep{Salimans2016}] --- A method aimed at hindering mode collapse, minibatch discrimination offers a method for penalizing strong similarity between generated samples. By adding a layer in the discriminator, minibatch discrimination is able to compute the cross-sample distance and measure the diversity of generated data. The discriminator still outputs a single probability, but offers the data computed with the use of minibatch as side information.
    
    \item[Historical averaging \citep{Salimans2016}] --- Designed to avoid the oscillating and cyclical behavior of GANs, historical averaging incorporates past values of parameters into the player's cost function by adding a term $\parallel\theta - \frac{1}{t}\sum_{i=1}^{t}\theta[i]\parallel^{2}$, with $\theta[i]$ being the value of the parameters at past time $i$.
    
    \item[One-sided label smoothing \citep{Salimans2016}] --- Hindering the discriminator from becoming overconfident and squashing generator's gradients, one-sided label smoothing restricts the discriminator to output the probabilities between 0.1 and 0.9, instead of 0 and 1, thus handicapping the discriminator and alleviating the vanishing gradient problem. 
    
    \item[Noise injection \citep{Arjovsky2017}] --- Designed to tackle the training instability problem, in particular, vanishing gradients problem, injecting continuous noise to the inputs of the discriminator leads to smoothing the distribution of the probability mass, thus reducing the problem of non-overlapping support, problematic especially for standard GAN. Noise injection has been shown to be computationally efficient and practical \citep{Jenni2019}, however, leading to a lower quality of the generated image \citep{Roth2017}.
\end{description}


\section{Open Problems}
\label{sec:openprob}

Based on the surveyed methods, this section highlights several open problems which may provide promising research directions towards stabilizing GAN training.

    \subsection{Combining Training Stabilization Methods}

With the number of approaches coming from different disciplines and perspectives proposed in recent years, it is unclear how these techniques align with each other and how combinations affect the overall performance. Thus, an interesting avenue of research is the investigation of how these methods could be potentially applied jointly. An example of this idea is BigGAN \citep{Brock2018}, which achieved state-of-the-art results by combining already-known methods and subsequently scaling them. A proper combination of training techniques, architectural changes, and loss function modifications could prove to yield superior results. This is especially important in the context of game theory and gradient-based GAN variants, which have been consistently omitted in the architectural and loss function GAN literature \citep{Hong2019, Wang2019, Creswell2018}. Although these type of GANs often posses significant limitations, the methods outlined by them such as fictitious play or no-regret algorithms could provide an additional advantage in stabilizing the GAN training procedure.

    \subsection{Actor-Critic Methods for GAN Training}

The link between GANs and reinforcement learning has already been described \citep{Hong2019, Pfau2017, Finn}; in particular, \citet{Pfau2017} discussed the connection between GANs and actor-critic methods. However, numerous possible connections remain unexplored. One of the potentially beneficial stabilization technique known to improve the stability in actor-critic methods is the use of replay buffers, enabling off-policy updates based on the action chosen. Although not directly applicable, an extension of replay buffers, prioritized experience replay \citep{Schaul2015} could serve as a conceptual inspiration for a different method for identifying and improving the networks on the outliers, with outliers being the samples with large discrepancy between real and generated samples.

    \subsection{Discriminator Generalization}

GAN training relies on the idea of the discriminator producing useful feedback to the generator. This, however, is often hard to attain. Furthermore, it is not entirely possible to evaluate the quality of the discriminator's feedback and performance solely on the basis of its loss \citep{Thanh-Tung2019}. For example, a theoretically robust discriminator could have in practice learnt to discern real from generated images only based on a small set of features with no proper generalization. A crucial aspect of the optimal discriminator is its ability to generalize to unseen samples. A proper GAN objective should induce the discriminator towards the optimal discriminator with good generalization capabilities. This issue has been explored more in-depth only recently, proposing new distances \citep{Off2018, Arora2017} and regularization techniques \citep{Thanh-Tung2019}. Nevertheless, constructing a GAN objective that fully incorporates the discriminator's generalization capabilities remains a challenge.

    \subsection{Opponent Modeling in GANs}

The two-player zero-sum game is a subject of rich literature in the field of opponent modeling \citep{Albrecht2018}. Nevertheless, very few studies have explored the connection between the GAN game and opponent modeling or multi-agent learning broadly. An interesting avenue would be to model the GAN game as a game between two agents with limited information and devise mechanisms that would allow to incorporate the awareness of the other agent into their training procedure. Such an approach could potentially lead to more stable convergence \citep{Schafer2019a}.

    \subsection{Variational Autoencoder GAN Hybrid}

Similar to GANs, Variational Autoencoders (VAEs) are also a type of generative model \citep{kingma2013auto}. However, in contrast to the former, which implicitly model a target distribution, VAEs follow an explicit approach to approximately estimating the data distribution through the use of an encoder and a decoder. VAEs are known to produce images of lesser quality in comparison to GANs. Nevertheless, they possess stronger theoretical groundings and have been shown to be less prone to mode collapse than GANs. Drawing on VAEs, VAEGAN \citep{Larsen2016} and $\alpha$-GAN \citep{Rosca2017} combined the approaches to alleviate the mode collapse problem. This area at the intersection of VAEs and GANs, although not yet entirely explored, provides promising results and could prove to achieve significant results, especially in terms of preventing mode collapse.

\section{Conclusion}
\label{sec:conc}

This article surveyed current approaches for stabilizing the GAN training procedure, categorizing and outlining various techniques and key concepts. We contributed a comparative summary of methods together with a description of potential research directions to address open problems. We conclude that although recent years have seen an increase of published studies providing a number of methods to overcoming GAN training instability, the majority of these studies are limited to preventing only one training issue and often lack rigorous theoretical justification.

\bibliography{gan-survey}

\end{document}